\documentclass[sigconf]{acmart}

\usepackage[utf8]{inputenc} 
\usepackage[T1]{fontenc}    
\usepackage{hyperref}       
\usepackage{amsfonts}       
\usepackage{microtype}      
\usepackage{graphicx}
\usepackage{subfigure}
\usepackage{multirow}
\usepackage{algorithm2e}
\usepackage{amsmath}

\usepackage{amssymb}
\usepackage{amsthm}
\usepackage{pgfplots}

\theoremstyle{plain}
\newtheorem{theorem}{Theorem}[section]

\theoremstyle{definition}

\theoremstyle{remark}

\pgfplotsset{compat=1.18}

\setcopyright{acmlicensed}
\copyrightyear{2026}
\acmYear{2026}
\setcopyright{cc}
\setcctype{by}
\acmConference[KDD 2026] {Proceedings of the 32nd ACM SIGKDD Conference on Knowledge Discovery and Data Mining V.1}{August 9--13, 2025}{Jeju Island, Republic of Korea.}
\acmBooktitle{Proceedings of the 32nd ACM SIGKDD Conference on Knowledge Discovery and Data Mining V.1 (KDD 2026), August 9--13, 2025, Jeju Island, Republic of Korea}
\acmISBN{979-8-4007-2258-5/2026/08}
\acmDOI{10.1145/3770854.3780179}

\begin{document}

\title{M2NO: An Efficient Multi-Resolution Operator Framework for Dynamic Multi-Scale PDE Solvers}

\author{Zhihao Li}
\orcid{0000-0003-4752-6811}
\affiliation{%
  \institution{The Hong Kong University of Science and Technology (Guangzhou)}
  \city{Guangzhou}
  \country{China}
}
\email{zli416@connect.hkust-gz.edu.cn}

\author{Zhilu Lai}
\affiliation{
    \institution{The Hong Kong University of Science and Technology (Guangzhou)}
    \city{Guangzhou}
    \country{China}\\
    \institution{The Hong Kong University of Science and Technology}
    \city{Hong Kong SAR}
    \country{China}
}
\email{zhilulai@ust.hk}

\author{Xiaobo Zhang}
\affiliation{
    \institution{Southwest Jiaotong University}
    \city{Chengdu}
    \country{China}
}
\email{zhangxb@swjtu.edu.cn}

\author{Wei Wang}
\authornote{Corresponding author.}
\affiliation{
    \institution{The Hong Kong University of Science and Technology (Guangzhou)}
    \city{Guangzhou}
    \country{China}\\
    \institution{The Hong Kong University of Science and Technology}
    \city{Hong Kong SAR}
    \country{China}
}
\email{weiwcs@ust.hk}

\renewcommand{\shortauthors}{Zhihao Li, Zhilu Lai, Xiaobo Zhang, \& Wei Wang}

\begin{abstract}
Solving high-dimensional partial differential equations (PDEs) efficiently requires handling multi-scale features across varying resolutions. To address this challenge, we present the Multiwavelet-based Multigrid Neural Operator (M2NO), a deep learning framework that integrates a multigrid structure with predefined multiwavelet spaces. M2NO leverages multi-resolution analysis to selectively transfer low-frequency error components to coarser grids while preserving high-frequency details at finer levels. This design enhances both accuracy and computational efficiency without introducing additional complexity. Moreover, M2NO serves as an effective preconditioner for iterative solvers, further accelerating convergence in large-scale PDE simulations. Through extensive evaluations on diverse PDE benchmarks, including high-resolution, super-resolution tasks, and preconditioning settings, M2NO consistently outperforms existing models. Its ability to efficiently capture fine-scale variations and large-scale structures makes it a robust and versatile solution for complex PDE simulations. Our code and datasets are available on \url{https://github.com/lizhihao2022/M2NO}.
\end{abstract}



\maketitle

\section{Introduction}
Partial differential equations (PDEs) underpin models in fluid dynamics, solid mechanics, and climate science \cite{82:intro,87:turb,20:turb,02:pde}.  Classical solvers, like finite–difference and finite–element methods, offer provable convergence but can be prohibitively expensive for high-dimensional, multi-scale problems \cite{12:fem}. Multigrid algorithms mitigate this cost by eliminating low-frequency errors on successively coarser grids, often reducing the required Krylov iterations by orders of magnitude \cite{00:tutorial,77:bvp}; however, they still rely on hand-crafted restriction/prolongation operators and demand problem-specific tuning for complex geometries. Recently, advancements in deep learning have introduced alternative approaches, such as Fourier Neural Operators (FNO) \cite{21:fno}, Galerkin Transformers \cite{21:GT}, and Deep Operator Networks (DeepONet) \cite{21:DeepONet}. These methods utilize neural network architectures to efficiently approximate mappings between parameters and solutions in infinite-dimensional spaces, tackling both forward and inverse problems in PDEs \cite{19:pinn,20:mgno,23:no,25:HarnessingSP}.

Yet existing neural operators face three persistent hurdles:
(i) their accuracy degrades when deployed on grids finer or coarser than those seen in training;
(ii) they struggle to capture both global low-frequency trends and sharp local features within a single network; and
(iii) they rarely match the iteration efficiency of hand-crafted multigrid solvers.
To overcome these limitations, we propose an innovative framework that integrates a multigrid structure within predefined multiwavelet spaces, significantly enhancing the precision and efficiency of learning PDE operators. 

Our methodology strategically transfers only the low-frequency error components to coarser grids for targeted elimination, drawing upon the robust principles of multigrid techniques \cite{00:tutorial,17:amg,00:amg,77:bvp}. Traditional multigrid methods, effective yet often limited by scalability challenges in very large or complex scenarios, are enhanced by our proactive selection of appropriate coarse spaces and specific restriction/prolongation operators. This ensures that our multigrid process scales efficiently, optimizing both computational resources and adapting to complex problem sizes. In more detail, we employ a low-pass filter as the restriction operator, which specifically transfers the low-frequency components of errors into coarser grids, and use its transpose for the prolongation phase. This low-pass filter is ingeniously constructed using multiwavelets basis, providing a compact and data-efficient representation of operators across various scales \cite{92:ten,00:Fourier,11:first,21:mwt}. The design process involves utilizing multiresolution analysis (MRA) to meticulously craft the multiwavelet transformation, integrating both the low-pass filter and its transpose as the fundamental restriction and prolongation operators \cite{93:wave, 96:schur, 07:fastwave}. This application of MRA within our framework significantly enhances our model’s ability to dynamically adapt to the evolving scales and complexities of PDE problems.

In this work, we introduce the Multiwavelet-based Multigrid Neural Operator (M2NO), a neural network architecture designed to inherently model multi-scale structures in PDEs, making notable contributions in the following areas:
\begin{itemize}
    \item \textbf{Multiresolution and Multiscale Representation:} We leverages MRA to design efficient restriction and prolongation operators based on multiwavelet transformations. By utilizing low-pass filters and their transposes, M2NO captures multi-scale information effectively, enabling precise mapping across different resolution levels. This structured representation allows for efficient error correction and improved adaptability in solving PDEs.

    \item \textbf{Enhanced Precision and Efficiency:} M2NO significantly enhances the accuracy and efficiency of solving dynamic multiscale PDEs through the integration of a multigrid structure within predefined multiwavelet spaces. This strategic alignment enables precise error management by selectively addressing low-frequency components, thus optimizing the solution processes across varied scales and complexities.

    \item \textbf{Comprehensive Empirical Validation:} M2NO has been rigorously evaluated across diverse 1D and 2D PDE benchmarks, including large-scale real-world datasets like ERA5, where it demonstrated strong performance in both multi-resolution and super-resolution tasks. Spectral Analysis further confirms M2NO's effectiveness across different frequency bands, highlighting its robustness in capturing both low- and high-frequency components. Additionally, M2NO's flexibility as a preconditioner for iterative solvers showcases its practical utility in accelerating convergence for complex PDE systems. These extensive experiments validate M2NO's scalability, accuracy, and versatility in high-dimensional scientific computing.
\end{itemize}

\section{Preliminaries}
In this section, we provide a concise overview of the multigrid method \cite{17:amg,00:tutorial} and multiresolution analysis using multiwavelets \cite{02:mw,92:ten}, highlighting their similarities and how they inform the development of M2NO.  

\subsection{Multigrid Method}
The multigrid method is a well-grounded numerical technique for solving systems of equations that approximate PDEs. Specifically, we address the problem of solving the linear equation system:
\begin{equation}
    A^{h}u^{h}=b^{h},
\end{equation}
where $A^{h}$ and $b^{h}$ are derived from the discretization of a differential equation on a grid $\Omega^{h}$, with $h$ representing the step size. 

Given a prolongation operator, $I_{2h}^{h}$, where the superscript denotes the fine grid and the subscript denotes the coarse grid, alongside a restriction operator, $I_{h}^{2h}$, we can recursively define a multigrid method. In the two-level V-cycle method, we first perform a few relaxation steps (typically one or two) on the fine grid $\Omega^{h}$ to obtain an initial estimate $u^{h}$. Subsequently, we calculate the residual $r^{h}=b^{h}-A^{h}u^{h}$; restrict this residual to the coarse grid $\Omega^{2h}$, yielding $\Omega^{2h}$, $r^{2h}=I_{h}^{2h}r^{h}$; and solve the error equation
\begin{equation}
    A^{2h}e^{2h}=r^{2h}
\end{equation}
on the coarse grid. After solving, update the solution on the fine grid by setting
\begin{equation}
    u^{h}=u^{h}+I_{2h}^{h}e^{2h}
\end{equation}
and perform a few more relaxation steps on the fine grid. This procedure outlines the foundational steps of the V-cycle multigrid scheme, which is recursively defined, as shown in Algorithm \ref{alg:v}. For a comprehensive discussion and theoretical grounding of the Multigrid methods, please refer to the Appendix \ref{appendix:amg}.

\subsection{Multiwavelets}
Wavelets are powerful tools for decomposing data, functions, or operators into different frequency components and analyzing them through scaling. They can form a complete orthonormal basis for \( L^2(\mathbb{R}) \). Due to their scaling properties, wavelet functions have time or space widths inversely related to their frequencies: they are narrow at high frequencies and broad at low frequencies. This characteristic ensures excellent localization in both the frequency domain and physical space, making wavelets particularly effective for analyzing phenomena across multiple scales.

A MRA using multiwavelets involves a sequence of closed subspaces, defined as:
\begin{equation*}
    \mathbf{V}_{0}^{k}\subset\mathbf{V}_{1}^{k}\subset\ldots\subset\mathbf{V}_{n}^{k}\subset\ldots
\end{equation*}
For $k=1,2,\ldots$, and $n=0,1,2,\ldots$, we define $\mathbf{V}_{n}^{k}$ as a space of piecewise polynomial functions with dimension $2^{n}k$, that satisfy certain conditions. The multiwavelet subspace $\mathbf{W}_{n}^{k}$, defined as the orthogonal complement of $\mathbf{V}_{n}^{k}$ in $\mathbf{V}_{n+1}^{k}$, satisfies:
\begin{equation}
    \mathbf{V}_{n}^{k}\oplus\mathbf{W}_{n}^{k}=\mathbf{V}_{n+1}^{k},\quad \mathbf{W}_{n}^{k}\bot\mathbf{V}_{n}^{k},
\end{equation}
and the dimension of $\mathbf{W}_{n}^{k}$ is $2^{n}k$. Therefore, the composition of spaces is:
\begin{equation}
    \mathbf{V}_{n}^{k}=\mathbf{V}_{0}^{k}\oplus\mathbf{W}_{0}^{k}\oplus\mathbf{W}_{1}^{k}\oplus\ldots\oplus\mathbf{W}_{n-1}^{k}.
\end{equation}

The operators $H$ and $G$, which transform the basis of the space $V_{j}$ (for $j\in\mathbb{Z}^{+}$) to the bases of the spaces $V_{j-1}$ and $W_{j-1}$ respectively, exhibit the following properties (assuming $H$ and $G$ are real-valued):
\begin{equation*}
    \begin{aligned}
        H^{T}H+G^{T}G&=I\\
        HG^{T}=GH^{T}&=0\\
        HH^{T}=GG^{T}&=I.
    \end{aligned}
\end{equation*}

In the two-dimensional discrete wavelet transform, depicted in Fig \ref{fig:dwt}(a), $H$ and $G$ function as low-pass and high-pass filters, respectively, allowing the partitioning of frequency components at each level. Fig \ref{fig:dwt}(b) visually represents this decomposition in a matrix format, enhancing our understanding of how submatrices $B_n$, $C_n$, and $D_n$ are derived from $L_n$. 

The wavelet transform is defined as $\mathcal{W}:V_{j}\to V_{j-1}\oplus W_{j-1}$ (for $j\in\mathbb{Z}^{+}$), and is given by:
\begin{equation}
    \mathcal{W}=\begin{pmatrix}
        H\\G
    \end{pmatrix}
\end{equation}

Owing to the orthogonality properties of $H$ and $G$, $\mathcal{W}$ maintains orthogonality. For detailed definitions of $H$, $G$, and $\mathcal{W}$ in 1D and 2D contexts, please refer to Appendix \ref{appendix:mw}.

\subsection{Similarities between Multigrid and MRA}
Our objective in comparing these two methodologies is to explore how each enhances the other, as detailed in the following points:
\begin{enumerate}
    \item The highest resolution space $V_{n}$ in MRA aligns with $\Omega^h$, which represents the space of the finest grid vectors in the multigrid.
    \item The decomposition in MRA, where data is broken down, corresponds to the restriction in multigrid. Similarly, the forward wavelet transform aligns with the restriction operator $I_{h}^{2h}$.
    \item The reconstruction in MRA, which reassembles the data, is equivalent to the prolongation in multigrid. Correspondingly, the inverse wavelet transform is equivalent to the prolongation operator $I_{2h}^{h}$.
    \item The orthogonality condition $\mathcal{W}\mathcal{W}^{T} = I$ corresponds to the identity relationship for the grid transfer operators $I_{h}^{2h}I_{2h}^{h} = I$.
    \item The multiresolution structure $V_{n}=V_{n-1}\oplus W_{n-1}$ in MRA parallels the multigrid structure $\Omega^{h}=\text{Range}\{ I_{2h}^{h} \}\oplus \text{Nullspace}\{ I_{h}^{2h} \}$.
\end{enumerate}

These comparisons highlight the synergy between the multigrid and multiresolution properties of wavelet transforms and the error smoothing and correcting capabilities of intergrid operators in multigrid methods.

\begin{figure*}
    \centering
    \includegraphics[width=\textwidth]{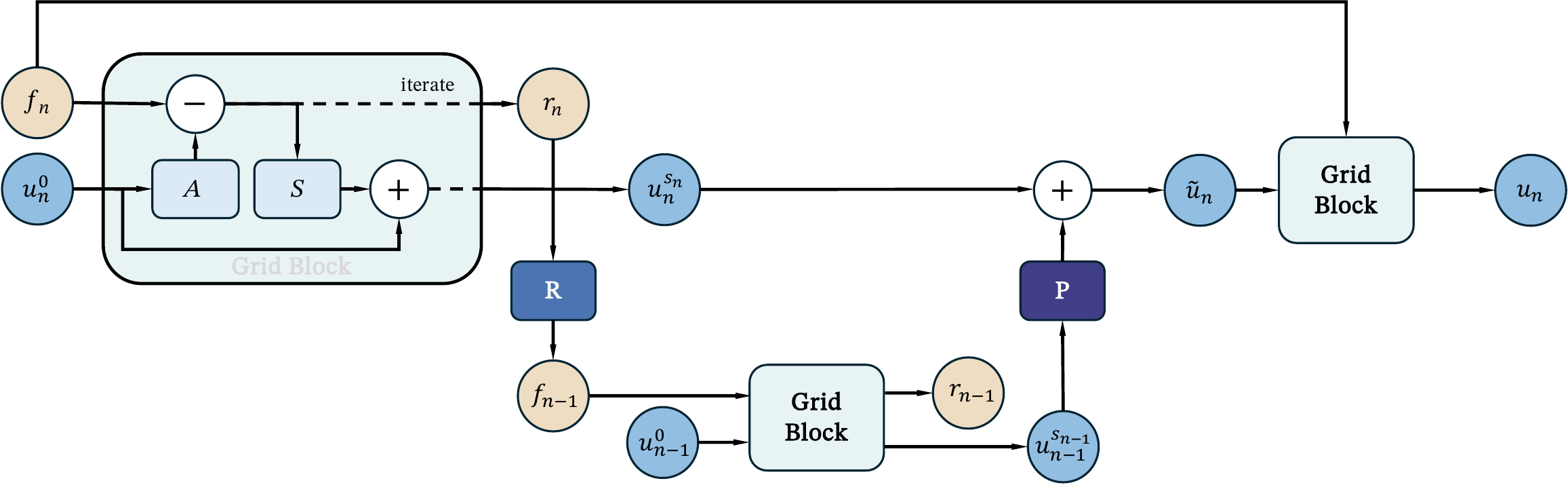}
    \vspace{-15pt}
    \caption{\textbf{Illustration of M2NO with Two Levels and Grid Block.} $f_{n}$ is initialized as $a_{n}$ and $u_{n}$ is initialized as $0$. $R$ and $P$ are restriction and prolongation operator respectively. $A$ and $S$ are implemented as single-layer convolutional neural networks.}
    \label{fig:Arch}
    \vspace{-10pt}
\end{figure*}

\section{Methodology}
This section explores the M2NO, leveraging the combined advantages of multigrid and multiresolution wavelet transformations. Initially, we define the problem and lay its theoretical groundwork in section \ref{sub:3-1}. Section \ref{sub:3-2} explains the design and practical implementation of multiwavelet-based operators essential for M2NO. Subsequently, section \ref{sub:3-3} illustrates these methodologies' application in solving discrete PDEs. 

\subsection{Problem Definition} \label{sub:3-1}
Formally, let $\mathcal{A}$ and $\mathcal{U}$ denote two Sobolev spaces $\mathcal{W}^{s,p}$ with $s>0$ and $p \geq 1$. In this context, the operator $\mathcal{G}$ maps from $\mathcal{A}$ to $\mathcal{U}$, $\mathcal{G}: \mathcal{A} \to \mathcal{U}$. We focus on cases where $s > 0$ and $p = 2$, choosing $p = 2$ specifically to enable the definition of projections within a Hilbert space structure $\mathcal{H}^{s}$, using measures $\mu$.

We define the operator $\mathcal{G}$ as an integral operator with a kernel $\kappa$, where $\kappa:D\times D\to L^{2}$. The action of $\mathcal{G}$ on an element $a \in \mathcal{A}$ is given by the integral equation:
\begin{equation}
    \mathcal{G}a(x)=\int_{D}{\kappa(x,y)a(y)dy}.
\end{equation}
This setup allows $\mathcal{G}$ to operate within the framework of Hilbert spaces, leveraging the properties of $L^{2}$ spaces to facilitate analysis and computations involving the operator.

An illustrative application of the theory described is the learning of the solution operator for a parametric PDE. Let $\mathcal{L}_{a}$ denote a differential operator that varies with a parameter $a\in \mathcal{A}$. Consider the general PDE defined as:
\begin{equation}
\label{eq:pde}
    (\mathcal{L}_{a}u)(x)=f(x),\quad x\in D,
\end{equation}
where $D$ is a bounded, open subset of $\mathbb{R}^d$, $f$ is a function residing in an appropriate function space, and the equation is subject to a boundary condition on $\partial D$. Assuming the PDE is well-posed, which implies there exists a unique solution under reasonable assumptions on the data and parameters, we define the operator $\mathcal{G}^\dagger: \mathcal{A} \to \mathcal{U}$. This operator maps the parameter $a$ to the solution $u$ of \eqref{eq:pde}, effectively encapsulating the dependency of the solution on the parameter within a functional mapping framework. 

\subsection{Multiwavelet-based Restriction and Prolongation Operators} \label{sub:3-2}
Given an orthonormal polynomial basis for \(V_{0}^{k}\) as \(\{\phi_{0}, \phi_{1}, \ldots, \phi_{k-1}\}\) with respect to the measure \(\mu_{0}\), we can generate the basis for subsequent spaces \(V_{n}^{k}\) for \(n>1\) through shift and scale (hence the multi-scale aspect) operations on the original basis:
\begin{equation}
    \phi_{jl}^{n} = 2^{\frac{n}{2}} \phi_{j}(2^n x - l),
\end{equation}
where $j=0,\ldots,k-1$, $l=0,\ldots,2^{n}-1$ with respect to \(\mu_{n}\), which is derived from the shift and scale transformations of \(\mu_{0}\). The filter coefficients construct the scaling function \(\phi\) and wavelet \(\psi\); in our context, functions \(\phi\) and \(\psi\) are known, and we utilize them to determine the filter coefficients \((H^{(0)}, H^{(1)}, G^{(0)}, G^{(1)})\), as they transform from a fine scale \(n=1\) to a coarser scale \(n=0\). A uniform measure \(\mu_{0}\) is discussed in \cite{02:mw}, and \cite{21:mwt} extends it to arbitrary measures by incorporating correction terms \(\sum^{(0)}\) and \(\sum^{(1)}\).

In one dimension, the operators \(H\) and \(G\) are constructed as
\begin{equation}
    H = \begin{pmatrix} H^{(0)} & H^{(1)} \end{pmatrix}, \quad
    G = \begin{pmatrix} G^{(0)} & G^{(1)} \end{pmatrix},
\end{equation}
to transform the basis of space \(V_{j}\) to the bases of spaces \(V_{j+1}\) and \(W_{j+1}\), respectively. In two dimensions, we utilize the tensor product of one-dimensional multiresolution analyses:
\begin{equation}
    H = H_{y} \otimes H_{x}, \quad
    G = \begin{pmatrix} G_{y} \otimes H_{x} \\ H_{y} \otimes G_{x} \\ G_{y} \otimes G_{x} \end{pmatrix},
\end{equation}
where \(\otimes\) denotes the Kronecker product. For detailed explanation, see Appendix \ref{appendix:mw}.

Our motivation is using the low-pass filter as a restriction operator for the multigrid. The restriction operator and prolongation operator are defined as 
\begin{equation} \label{eq:lpf-r}
    I_{h}^{2h} = H,
\end{equation}
\begin{equation} \label{eq:lpf-p}
    I_{2h}^h=(I_{h}^{2h})^{T}=H^{T}.
\end{equation}

At each decomposition level, it is predominantly the low-pass filters that encapsulate the approximation of the original matrix as illustrated in each one of the levels of scale in Fig \ref{fig:dwt}. In the majority of cases, this methodology proves functional because the residual error associated with the matrices $B$, $C$, and $D$ is almost negligible. 

With the explicit computation of the low-pass filter \(H\), we take the restriction operator as \(I_{h}^{2h} = H\) and the corresponding prolongation operator as \(I_{2h}^{h} = H^T\). It is noteworthy that the prolongation operator is identical for both spaces \(V_{0}\) and \(V_{1}\), which correspond to the coarsest and the next finer grids, respectively.

\begin{table*}[t]
    \caption{\textbf{Performance comparison with baselines on benchmarks.} $L_{2}$ loss is recorded.}
    \label{table:all}
    \vspace{-8pt}
    \begin{sc}
        \renewcommand\arraystretch{0.85}
        \renewcommand{\multirowsetup}{\centering}
        \footnotesize 
        \resizebox{\linewidth}{!}{
        \begin{tabular}{c|cc|ccc|c}
            \toprule
                Model & Advection & Burgers & Darcy & NS & Diff-React & ERA5 (Temp) \\ 
            \midrule
                UNet & 3.44E-05 & 3.83E-05 & 1.54E-01 & 2.15E-02 & 1.46E-01 & 7.55E-01\\
                FNO & 1.68E-05 & \underline{1.81E-05} & 1.95E-02 & 1.72E-02 & 1.64E-01 & 4.06E-03 \\
                MgNO & /  & / & 5.72E-02  & 1.27E-02 & \underline{1.12E-01} & 9.76E-01 \\
                MWT & \underline{9.91E-06} & 2.07E-05 & 1.91E-02  & \underline{1.19E-02} & 1.15E-01 & 5.28E-03 \\
            \midrule
                LSM & 3.44E-02 & 1.51E-02 & 4.69E-02 & 1.61E-02 & 1.58E-01 & 4.16E-03\\ 
                GNOT & 1.21E-02 & 1.08E-02 & 2.14E-02 & 4.57E-02 & 1.47E-02 & 5.03E-03 \\
                ONO & 1.23E-02 & 1.17E-02 & 2.08E-02 & 3.93E-02 & 1.52E-02 & 4.13E-03 \\
                Transolver & 1.01E-02 & 1.09E-02 & \underline{1.73E-02} & 3.65E-02 & 1.52E-02 & 4.41E-03\\
                AMG & 8.89E-03 & 1.05E-02 & \underline{1.73E-02} & 3.42E-02 & 1.40E-02 & / \\ 
            \midrule
                \textbf{M2NO} & \textbf{2.35E-06} & \textbf{7.52E-06} & \textbf{9.64E-03} & \textbf{3.61E-03} & \textbf{1.01E-01} & \textbf{2.77E-03} \\
                Error Reduction & 78.41\% & 60.49\% & 49.56\% & 69.64\% & 9.82\% & 31.69\% \\ 
            \bottomrule
        \end{tabular}}
    \end{sc}
\end{table*}

\subsection{Multiwavelet-based Multigrid Method for Discrete PDEs} \label{sub:3-3}
M2NO is a surrogate neural operator maps from the linear finite element space of input functions $a\in\mathcal{A}$, to the linear finite element space of output functions $u\in\mathcal{U}$. The mapping is defined as:
\begin{equation}
    \begin{cases}
        h^{0} = a \in \mathcal{A},\\
        h^{l}(a) = \sigma(\mathcal{L}_{mg}^{l}h^{l-1}(a) + B^{l}h^{l-1}(a)+b^{l}),&l=1:L\\
        u = \mathcal{G}_{\theta}(a) = \mathcal{L}_{mg}^{L+1}(h^{L}(a))\in\mathcal{U},
    \end{cases}
\end{equation}
where $\sigma$ denotes the point-wise GELU activation function, and $\mathcal{L}^{\ell}_{mg}$, $B^{l}$, and $b^{l}$ are parameters of the linear operators at each layer.

For the multigrid correction particularly, we employ a linear Finite Element Method (FEM) discretization with mesh size $h=\frac{1}{d}$, the discretized system is described by:
\begin{equation}
\label{ls}
    A_{j} * u_{j} = f_{j},
\end{equation}
where $u_{j}, f_{j} \in \mathbb{R}^{d \times d}$. In this model, $*$ denotes the convolution operation for a single channel, using a $3 \times 3$ kernel $A$ determined by the elliptic operator and specific boundary conditions. The inverse of $A *$ aligns with the discrete Green's function in a linear FEM context. The V-cycle multigrid approach, which can be represented as a conventional convolutional neural network, effectively solves this equation \cite{23:interp, 19:mgnet}.

We provide a concise overview of the essential components of the multigrid structure as shown in Fig \ref{fig:Arch}, framed in the language of convolution as an operator mapping from $f$ to $u$:
\begin{enumerate}
    \item \textbf{Initialization:} Set the initial function $f_{n}$ as $a_{n}$ and the initial guess $u_{n}^{0} (j=0,\ldots,n)$.
    \item \textbf{Pre-smoothing:} Apply $s_{n}$ steps of the smoothing iteration $S_{n}$ to obtain an approximation:
    \begin{equation}
        u_{n}^{i} = u_{n}^{i-1} + S_{n}^{i} * (f_{n} - A_{n} * u_{n}^{i-1}), \quad i=1,\ldots,s_{n},
    \end{equation}
    with residual error defined as $r_{j} = f_{n} - A_{n} * u_{n}^{i-1}$.
    \item \textbf{Multigrid correction:} Transition to a coarser grid, where a coarse grid operator $A_{j-1}$ (for $j=1,\ldots,n$) is defined. The residual $r_{j} = f_{j} - A_{j} * u_{j}^{s_{j}}$ is restricted using a multiwavelet-based restriction operator $I_{h}^{2h}$, defined in \eqref{eq:lpf-r}. The correction equation \( A_{j-1} e_{j-1} = r_{j-1} \) is solved through several relaxation iterations denoted as \( s_{j-1} \). In this equation, \( e_{j-1} \) represents the error of the linear system; for ease of notation, we refer to it as \( u_{j-1} \) throughout the paper. The result, $u_{j-1}$, is prolonged back to the finer grid using the multiwavelet-based prolongation operator $I_{2h}^{h}$, as defined in \eqref{eq:lpf-p}, to update the approximation:
    \begin{align*}
        f_{j-1} &= I_{h}^{2h}r_{j}, \\
        u_{j-1}^{i} &= u_{j-1}^{i-1} + S_{j-1}^{i} * (f_{j-1} - A_{j-1} * u_{j-1}^{i-1}), \\
        \tilde{u}_{j} &= u_{j}^{s_{j}} + I_{2h}^{h} u_{j-1}^{s_{j-1}},
    \end{align*}
    where $i=1,\ldots,s_{j-1}$, is the iterative number in each level.
    \item \textbf{Post-processing:} At this stage, one may choose to further prolong to a finer level in a process known as super-resolution or continue with post-smoothing steps.
\end{enumerate}
We then give the convergence of the operators within the framework, detailed proof of this can be found in Appendix \ref{appendix:proof}.
\begin{theorem}[Approximation of Continuous Operators by M2NO] \label{th:conv}
Let $X = H^s(\Omega)$ and $Y = H^{s'}(\Omega)$ for some $s, s' \geq 1$, and let $\sigma \in C(\mathbb{R})$ be a non-polynomial activation function used in the M2NO architecture. Given any continuous operator $\mathcal{G}^*: X \to Y$, a compact set $C \subseteq X$, and a desired accuracy $\epsilon > 0$, there exists a neural network configuration within the M2NO framework, specifically a network with $n$ neurons per layer, such that
\begin{equation}
    \inf_{\mathcal{G} \in \mathcal{N}_n} \sup_{u \in C} \|\mathcal{G}^*(u) - \mathcal{G}(u)\|_Y \leq \epsilon,
\end{equation}
where $\mathcal{N}_n$ denotes the set of M2NO. 
\end{theorem}

\subsection{Model Implementation}
In the M2NO, the neural networks $A$, $S$, $B$, $C$ and $D^{-1}$ are implemented as single-layer convolutional neural networks (CNNs). The orthonormal polynomial (OP) basis utilized is the Legendre (Leg) basis, chosen with both uniform and non-uniform measures $\mu_{0}$, in alignment with the MWT\cite{21:mwt} methodologies. 

From a computational perspective, the restriction and prolongation operators depicted in Eq \eqref{eq:lpf-r}-\eqref{eq:lpf-p} and Fig \ref{fig:Arch} specifically target the transformation of multiscale and multiwavelet coefficients. Despite this focus on coefficients, the actual input and output of the model are point-wise function samples, specifically $(a_{i},u_{i})$. To address this, we synthesize the data sequence into conceptual functions $f_{a} = \sum_{i=1}^{N}a_{i}\phi_{ji}^{n}$ and $f_{u} = \sum_{i=1}^{N}u_{i}\phi_{ji}^{n}$, positing them within the space $V_{n}^{k}$ where $n=\log_{2}N$. It is critical to note that these functions, $f_{a}$ and $f_{u}$, are employed primarily for theoretical articulation and not for direct computation, serving as a notational convention rather than functional entities in the model's operations.

\textbf{Choosing Multiwavelet Basis.}  We have experimented with multiple wavelet bases, including Haar wavelets, Daubechies 4-tap wavelets, and multi-wavelets derived from Legendre and Chebyshev polynomials, all of which have demonstrated effectiveness in PDE-related applications. After careful evaluation, we selected Legendre polynomial-based multi-wavelets due to their superior empirical performance and compatibility with our M2NO architecture. The Legendre basis is particularly advantageous because its orthogonality and polynomial approximation properties provide an efficient, compact representation of PDE operators, significantly improving the model's ability to capture both global and local solution features. Additionally, we offer Chebyshev polynomial-based multi-wavelets as an optional hyperparameter choice, which also consistently outperform previous methods, underscoring the robustness and flexibility of our approach.

\section{Experiments} \label{sec:exp}
In this section, we evaluate the performance of the M2NO across a range of PDE benchmarks that incorporate various boundary conditions and conduct Multiresolution Analysis. Additionally, we delve into Spectral Analysis to assess M2NO's effectiveness in managing frequency-based discrepancies and explore its capabilities as a Preconditioner.

\textbf{General Setting.} We consider a set of function pairs $(a_i, u_i)_{i=1}^{N}$, where $a_i = a(x_i)$ and $u_i = u(x_i)$ at discretized points $x_i$ in the domain $D$. We express the functions $f_a$ and $f_u$ as linear combinations of basis functions $\phi_{ji}^{n}$, formulated as $f_a = \sum_{i=1}^{N} a_i \phi_{ji}^{n}$ and $f_u = \sum_{i=1}^{N} u_i \phi_{ji}^{n}$, both of which belong to the space $V_n^k$ where $n = \log_2 N$. These formulations serve as notational conventions rather than explicit constructs. Given the dataset $(a_i, u_i)_{i=1}^{N}$, our goal is to approximate the operator $\mathcal{G}$ by optimizing network parameters $\theta$. This involves solving the following optimization problem:
\begin{equation}
    \min_{\theta \in \Theta} \mathcal{L}(\theta) := \min_{\theta \in \Theta} \frac{1}{N} \sum_{i=1}^{N} \left[ \lVert \tilde{\mathcal{G}}_{\theta}(a_i) - u_i \rVert^2 \right].
\end{equation}

\textbf{Benchmarks.} To substantiate the superiority of our method, we conducted comprehensive method comparisons across multiple benchmark scenarios \cite{21:fno,22:bench}.  For Advection, Burgers and Darcy datasets, the split is 60\% for training, 20\% for validation, and 20\% for testing. For the Diffusion-Reaction and Navier-Stokes datasets, the data is allocated with 80\% used for training, 10\% for validation, and 10\% for testing. For ERA5 dataset, we use 12 days for training, 4 days for validation and 4 days for testing.

\textbf{Baselines.} We conducted a comprehensive comparison against established methods including UNet \cite{15:unet}, Fourier Neural Operator (FNO) \cite{21:fno}, Multiwavelet Transform (MWT) \cite{21:mwt}, and Multigrid Neural Operator (MgNO) \cite{23:mgno}. We also included transformer-based methods Latent Spectral Model (LSM) \cite{23:LSM}, Transolver \cite{24:Transolver}, GNOT \cite{23:GNOT}, ONO \cite{24:ImprovedOL} and AMG \cite{25:HarnessingSP}. All baselines were reproduced from their official repositories, and for each we performed a light hyper-parameter sweep in the vicinity of the authors’ published configurations to report the strongest reproducible numbers. Transolver and LSM were originally released only for 2-D grids; following the AMG \cite{25:HarnessingSP} setting, we treat the 1-D domain as an \emph{unstructured point cloud} so these models can be evaluated on the 1-D benchmarks. To ensure fairness, every model on a given dataset is trained under an identical optimisation protocol: 500 epochs, mean-squared ($L_{2}$) loss \cite{21:fno}, ADAM optimiser \cite{14:adam} with initial learning rate $10^{-3}$ decayed by a factor $\gamma=0.5$ every 100 epochs, and batch size 512 (1-D) or 16 (2-D). Hardware settings are also unified: Darcy and ERA5 runs use a single NVIDIA A6000 (48 GB), while all other experiments run on a single NVIDIA RTX 4090 (24 GB). Because of the very high resolution of ERA5, several large Transformer-based baselines could not be trained with their original widths—or in some cases could not finish training—within the available memory budget; reported ERA5 numbers therefore correspond to the largest configuration that fit on the GPU.

\textbf{Time-dependent tasks.} In our benchmarks, all tasks are time-dependent, except for Darcy Flow. These tasks are auto-regressive, requiring sequential predictions for each subsequent time step. Accordingly, we employ the 2D versions of the Multiwavelet-based Multigrid Neural Operator (M2NO), along with the Fourier Neural Operator (FNO) and Multiwavelet Transform (MWT) models, to predict the next time frame in a step-by-step manner across all benchmarks.

\subsection{Main Results} \label{sec:res}
Table \ref{table:all} presents the performance comparison of M2NO against baseline methods on a diverse set of PDE benchmarks, including Advection, Burgers, Darcy, Navier-Stokes (NS), Diffusion-Reaction (Diff-React), and the large-scale climate dataset ERA5. M2NO achieves the lowest $L_2$ loss across all tasks, with notable improvements of up to \textbf{78.41\%} on Advection and \textbf{69.64\%} on Navier-Stokes over the next best baseline. For high-resolution datasets like ERA5, M2NO surpasses Transolver, the closest competitor, with a margin of \textbf{31.69\%} in $L_2$ loss, demonstrating its robustness in capturing multi-scale features effectively. These results highlight M2NO's superior ability to manage both low- and high-frequency error components, providing consistent and scalable accuracy across different PDE types and resolutions. Detailed spectral analysis and multi-resolution studies further substantiate these improvements, as shown in the following sections.

\textbf{Real-World Data.} Table~\ref{table:era5} presents the quantitative results on the ERA5 test set, across five prediction targets: temperature (L1 and L2 error), wind U, wind V, and velocity magnitude (L2 error). M2NO achieves the lowest error across all metrics. Notably, for temperature prediction, M2NO reduces the L2 error by 31.69\% compared to the next best baseline (LSM), and achieves a similar 33.92\% improvement in L1 error, highlighting its superiority in capturing both average and extreme values. For wind components, M2NO achieves 17.26\% and 14.60\% lower L2 error than the best baseline, demonstrating robust modeling of fine-scale atmospheric flows. In velocity magnitude, M2NO shows a 17.18\% reduction in error, further illustrating its ability to resolve complex vector fields. These consistent improvements indicate that M2NO is more effective in learning multi-scale dynamics and reconstructing high-resolution physical fields from sparse or noisy data. 

\begin{table}[h]
\centering
\caption{\textbf{Performance comparison with baselines on ERA5}. We evaluate four physical variables including temperature, wind components (U, V), and velocity magnitude.}
\label{table:era5}
\vspace{-5pt}
\begin{sc}
\resizebox{\linewidth}{!}{
\begin{tabular}{c|c|cccc}
\toprule
\multirow{2}{*}{Model} & \multicolumn{2}{c}{Temperature} & Wind U & Wind V & Velocity \\ 
\cmidrule(lr){2-2} \cmidrule(lr){3-6} & L1 & \multicolumn{4}{c}{L2} \\
\midrule
Unet        & 7.21E-02 & 8.31E-02 & 7.60E-02 & 2.32E-01 & 7.55E-01 \\
FNO         & 3.04E-03 & 4.06E-03 & 1.13E-01 & 3.88E-01 & 7.85E-01 \\
MGNO        & 9.76E-01 & 9.76E-01 & 1.83E-01 & 2.43E-01 & 7.79E-01 \\
MWT         & 3.63E-03 & 5.28E-03 & 1.19E-01 & 3.72E-01 & 7.87E-01 \\
LSM         & 2.80E-03 & 4.16E-03 & 9.55E-02 & 2.43E-01 & 7.55E-01 \\ 
Transolver  & 3.06E-03 & 4.41E-03 & 9.90E-02 & 2.42E-01 & 7.87E-01 \\
\midrule
\textbf{M2NO} & \textbf{1.85E-03} & \textbf{2.77E-03} & \textbf{6.29E-02} & \textbf{1.98E-01} & \textbf{6.25E-01} \\
Error Reduction & 33.92\% & 31.69\% & 17.26\% & 14.60\% & 17.18\% \\ 
\bottomrule
\end{tabular}}
\end{sc}
\end{table}

\begin{figure*}[t]
    \centering
    \includegraphics[width=0.8\textwidth]{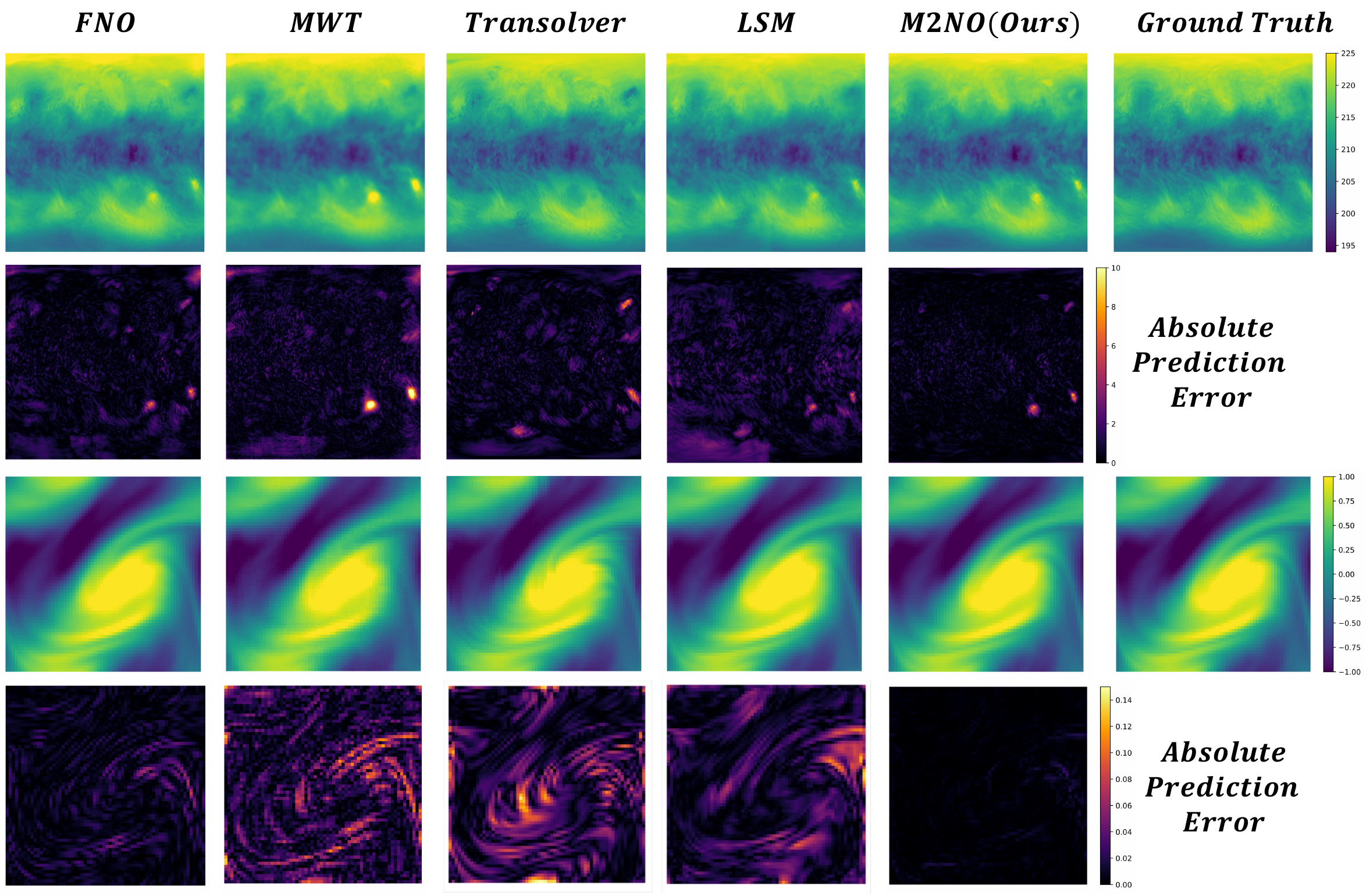}
    \caption{Performance showcases on the ERA5 (above) and NS (below).}
    \label{fig:showcase}
\end{figure*}


\textbf{Showcases.} Additional qualitative evaluation is provided in Figure \ref{fig:showcase}. M2NO achieves significantly lower absolute prediction errors compared to baselines on both ERA5 and NS datasets, capturing finer details and complex structures more accurately. The visualizations clearly demonstrate M2NO’s improved capability to model intricate spatial features inherent in multi-scale and high-resolution PDE solutions.

\subsection{Multiresolution Analysis} \label{sec:mra}

\begin{table}[ht]
\vspace{-5pt}
\centering
\caption{L2 errors of different models on the ERA5 across increasing spatial resolutions.}
\label{table:era5_res}
\vspace{-5pt}
\begin{sc}
\resizebox{\linewidth}{!}{
\begin{tabular}{c|ccccc}
\toprule
Model & 32$\times$32 & 64$\times$64 & 128$\times$128 & 256$\times$256 & 512$\times$512 \\ 
\midrule
UNet & 1.33E-01 & 9.85E-02 & 8.94E-02 & 8.54E-02 & 8.31E-02 \\
FNO & 4.28E-03 & 4.10E-03 & 4.23E-03 & 4.14E-03 & 4.06E-03 \\
MgNO & 1.52E-02 & 5.24E-01 & 9.01E-01 & 9.76E-01 & 9.76E-01 \\
MWT & 8.09E-03 & 5.46E-03 & 5.05E-03 & 5.04E-03 & 5.28E-03 \\
LSM & 4.89E-03 & 3.97E-03 & 4.42E-03 & 4.82E-03 & 4.16E-03 \\
Transolver & 4.03E-03 & 3.95E-03 & 4.03E-03 & 3.85E-03 & 4.41E-03 \\
\midrule
\textbf{M2NO} & \textbf{3.73E-03} & \textbf{3.68E-03} & \textbf{3.24E-03} & \textbf{2.88E-03} & \textbf{2.77E-03} \\
Error Reduction & 7.59\% & 6.64\% & 19.54\% & 25.11\% & 31.69\% \\
\bottomrule
\end{tabular}}
\end{sc}
\vspace{-5pt}
\end{table}

\textbf{Performance Across Multiple Resolutions.}
Table~\ref{table:era5_res} presents the $L_2$ errors of different models on the ERA5 dataset under varying spatial resolutions, ranging from $32\times32$ to $512\times512$. M2NO consistently achieves the lowest error across all resolutions, demonstrating its robustness and adaptability in capturing complex multi-scale patterns without the need for parameter adjustments. Notably, the improvement margin widens as resolution increases, reaching a significant 31.69\% promotion over the best baseline at $512\times512$.

\begin{table*}[ht]
\centering
\caption{\textbf{Computational efficiency comparison across models} (measured with input size 512$\times$512$\times$3, batch size 16).}
\vspace{-8pt}
\label{table:efficiency}
\begin{sc}
\resizebox{\linewidth}{!}{
\begin{tabular}{l|cc|ccc}
\toprule
Model & Param Count & Param (MB) & GPU Mem (MiB) & Train (s/epoch) & Inference (s/epoch) \\
\midrule
UNet            & 31,037,633  & 118.44 & 29,738 & 13.31 & 1.45 \\
FNO             & 1,444,017   & 10.97  & 15,740 & 5.36  & 0.87 \\
MgNO            & 227,403,798 & 867.47 & 43,768 & 42.73 & 3.34 \\
MWT             & 5,089,281   & 37.47  & 23,480 & 20.77 & 2.16 \\
LSM             & 4,801,089   & 18.33  & 42,930 & 62.54 & 2.99 \\
Transolver      & 279,607     & 1.06   & 41,450 & 25.54 & 3.14 \\
\midrule
\textbf{M2NO (Ours)} & \textbf{417,377} & \textbf{1.59} & \textbf{17,992} & \textbf{18.39} & \textbf{1.78} \\
\bottomrule
\end{tabular}}
\end{sc}
\end{table*}

\begin{table}[h]
\vspace{-5pt}
\centering
\caption{Super-resolution benchmarks on ERA5: $L_2$ loss across resolutions trained at $32\times 32$.}
\label{table:era5_super_resolution}
\vspace{-5pt}
\begin{sc}
\resizebox{\linewidth}{!}{
\begin{tabular}{c|cccc}
\toprule
Model & 64$\times$64 & 128$\times$128 & 256$\times$256 & 512$\times$512 \\ 
\midrule
FNO & 1.77E-01 & 3.20E-01 & 3.29E-01 & 3.33E-01 \\
MWT & 2.86E-01 & 3.39E-01 & 5.00E-01 & 6.64E-01 \\
LSM & 2.74E-02 & 4.69E-02 & 6.17E-02 & 6.84E-02 \\
\midrule
\textbf{M2NO} & \textbf{1.72E-02} & \textbf{2.28E-02} & \textbf{2.52E-02} & \textbf{2.63E-02} \\
Error Reduction & 37.10\% & 51.42\% & 59.14\% & 61.52\% \\
\bottomrule
\end{tabular}}
\end{sc}
\vspace{-5pt}
\end{table}

\textbf{Super-Resolution Performance.}  
Table \ref{table:era5_super_resolution} presents the super-resolution benchmarks for ERA5, with models trained at a base resolution of \(32 \times 32\) and tested across higher resolutions. M2NO consistently achieves the lowest $L_2$ losses at each upscale step, outperforming FNO, MWT, and LSM by significant margins. Notably, M2NO's error reduction is more pronounced as the resolution increases, achieving over \textbf{60\% improvement} at \(512 \times 512\). This highlights M2NO's robust capacity to recover fine-scale details in super-resolution tasks, benefiting from its multiwavelet-based multigrid architecture to effectively capture both local and global structures.

\subsection{Spectral Analysis}
We evaluate M2NO's ability to reduce high-frequency errors compared to other neural operator models using a two-dimensional Fast Fourier Transform (FFT) to analyze the radial energy spectrum (Fig. \ref{fig:spectral}). The results show that while all models struggle with low-frequency components, M2NO consistently achieves lower error energy across both low and high frequencies. This reflects its strength in capturing large-scale structures and resolving fine-scale details, making it particularly effective for precision-demanding tasks like Navier-Stokes simulations.

\definecolor{myred}{RGB}{219,049,036}
\definecolor{myorange}{RGB}{252,140,090}
\definecolor{myblue}{RGB}{075,116,178}
\definecolor{mywater}{RGB}{144,190,224}
\definecolor{myyellow}{RGB}{255,223,146}
\definecolor{mygrey}{RGB}{230,241,243}

\definecolor{vir0}{RGB}{068,004,090}
\definecolor{vir1}{RGB}{065,062,133}
\definecolor{vir2}{RGB}{048,104,141}
\definecolor{vir3}{RGB}{031,146,139}
\definecolor{vir4}{RGB}{053,183,119}
\definecolor{vir5}{RGB}{145,213,066}
\definecolor{vir6}{RGB}{248,230,032}

\pgfplotsset{width=0.86\linewidth,height=4cm,scale only axis}
\pgfplotsset{every axis/.append style={
font=\small,
line width=1.5pt,
tick style={line width=0.5pt}}}
\begin{figure}[ht]
    \centering
    \begin{tikzpicture}
    \begin{axis}[
    legend style={at={(0.96,0.96)},anchor=north east,legend columns=2, draw=none},
    xlabel=\textsc{Frequency},
    ylabel=\textsc{Average Error Energy},
    ylabel style={yshift=-0.5em},
    xtick={0,10,20,30,40},
    ymajorgrids=true,
    ymode=log,
    grid style=dashed,
    colormap={greenyellow}{rgb255(0cm)=(0,128,0); rgb255(1cm)=(255,255,0)},
    ]
    \addplot[mark=diamond,color=vir6] coordinates {(0,33.384992) (1,9.66359144) (2,6.16340039) (3,4.35228087) (4,3.89715445) (5,3.1641429) (6,2.46385979) (7,1.93984899) (8,1.96530872) (9,1.64540642) (10,1.80455963) (11,1.72066085) (12,1.62541929) (13,1.29138117) (14,1.31013305) (15,1.28585058) (16,1.32819245) (17,1.34942555) (18,1.1892619) (19,1.07043677) (20,1.04451125) (21,1.02569947) (22,0.99070156) (23,0.94219927) (24,0.86300129) (25,0.8262443) (26,0.87729947) (27,0.85598596) (28,0.94872755) (29,1.08940512) (30,1.01218248) (31,0.99378487) (32,0.92348119) (33,0.9748795) (34,1.03955396) (35,0.97935929) (36,1.0265305) (37,0.95243701) (38,0.99687879) (39,1.05029319) (40,0.92986704) (41,1.00508075) (42,0.85956513) (43,1.15632949) (44,1.90118812)};
    \addplot[mark=+,color=vir5] coordinates{(0,33.38714993) (1,7.17309378) (2,4.74260334) (3,3.41425217) (4,3.28882999) (5,3.08264548) (6,2.78735882) (7,2.12300182) (8,2.13973491) (9,1.85268939) (10,2.0480934) (11,1.93379721) (12,1.60838987) (13,1.5721706) (14,1.53576449) (15,1.4207633) (16,1.24732197) (17,1.25133989) (18,1.22206629) (19,1.18708632) (20,1.08975927) (21,0.97326592) (22,0.9299926) (23,0.84673782) (24,0.86571002) (25,0.83149241) (26,0.84925955) (27,0.83185313) (28,0.84257938) (29,0.831307) (30,0.81680459) (31,0.82709896) (32,0.7802005) (33,0.8089689) (34,0.89871688) (35,0.99396392) (36,0.93826532) (37,0.88238405) (38,0.85799163) (39,0.76702938) (40,0.69754801) (41,0.85902516) (42,0.89519886) (43,0.74935811) (44,0.88269566)};
    \addplot[mark=star,color=vir3] coordinates{(0,25.20139408) (1,6.01634103) (2,3.66574503) (3,3.38219722) (4,2.88274992) (5,2.01334836) (6,1.91101501) (7,1.53730737) (8,1.52683459) (9,1.36127258) (10,1.40582832) (11,1.3319479) (12,1.20839946) (13,1.06949847) (14,1.01253066) (15,0.97672967) (16,0.9302939) (17,0.95471261) (18,0.90160598) (19,0.8317993) (20,0.83816485) (21,0.81742671) (22,0.71112079) (23,0.71271059) (24,0.69851016) (25,0.66372345) (26,0.74827609) (27,0.69048196) (28,0.68857442) (29,0.66439498) (30,0.60021442) (31,0.70715498) (32,0.63893) (33,0.6654221) (34,0.65158827) (35,0.64446835) (36,0.56618973) (37,0.57245785) (38,0.52744572) (39,0.56049761) (40,0.59789652) (41,0.65739478) (42,0.80329412) (43,0.75679058) (44,0.68472163)};
    \addplot[mark=triangle,color=vir0] coordinates{(0,11.27271593) (1,3.65135503) (2,2.74461395) (3,2.04805807) (4,1.36259617) (5,0.95946919) (6,0.78962158) (7,0.72215751) (8,0.64767872) (9,0.59100537) (10,0.55881238) (11,0.52675399) (12,0.50745959) (13,0.50169103) (14,0.44991908) (15,0.47303842) (16,0.43893806) (17,0.43601245) (18,0.46685591) (19,0.45641545) (20,0.46461168) (21,0.40667106) (22,0.3782349) (23,0.38531928) (24,0.3867222) (25,0.36400572) (26,0.35414368) (27,0.3378821) (28,0.33531984) (29,0.32831732) (30,0.34287972) (31,0.35717593) (32,0.33908761) (33,0.31574747) (34,0.33316822) (35,0.29580023) (36,0.28899806) (37,0.31955271) (38,0.28085135) (39,0.2510116) (40,0.29578599) (41,0.23672594) (42,0.25349051) (43,0.2117891) (44,0.26215256)};
    \draw[dashed, thick, gray] (axis cs:10,1e-2) -- (axis cs:10,1e2);
    \legend{MGNO,FNO,MWT,M2NO}
    \end{axis}
    \end{tikzpicture}
    \vspace{-10pt}
    \caption{Models' Error Energy Spectrum on the NS.}
    \label{fig:spectral}
    \vspace{-10pt}
\end{figure}

\subsection{M2NO as Preconditioner}

We evaluate the effectiveness of M2NO as a preconditioner within the GMRES solver for solving a 512-dimensional Poisson equation. Specifically, we use the trained M2NO model to approximate the inverse operator by providing an identity matrix $\mathbf{I}$ as input and utilizing the output $\mathcal{G}(\mathbf{I})$ as the preconditioning matrix $\mathbf{M}$ \cite{25:NPO}. This learned preconditioner is then integrated into the GMRES solver to improve convergence. Baseline methods, including traditional numerical preconditioners (Gauss-Seidel, Schwarz) and other neural operator models (UNet, FNO, Transolver), are similarly integrated into GMRES for direct comparison.

Figure \ref{fig:preconditioner} compares the convergence behaviors across these methods. M2NO achieves residuals below $10^{-11}$ in approximately $50$ iterations, significantly outperforming all baselines. Traditional preconditioners require substantially more iterations to reach comparable residuals, and neural baselines converge more slowly and plateau at higher residual levels. These results highlight M2NO’s capability to provide efficient and accurate preconditioning through its multiwavelet-based multigrid structure, significantly enhancing solver efficiency for complex, high-dimensional PDEs.

\begin{figure}[h]
\begin{tikzpicture}
\begin{axis}[
    xlabel=\textsc{Iteration},
    ylabel=\textsc{Relative Residuals},
    ylabel style={yshift=-0.8em},
    ymode=log,
    grid=major,
    xmin=0,
    legend style={at={(0.5,1.0)}, anchor=south, legend columns=3, font=\small, draw=none},
]
    
    \addplot[color=vir1, mark=star] coordinates {
        (0, 1.0) (10, 0.9419617275661877) (20, 0.9240541787567774) (30, 0.8995001266856225) (40, 0.8746301582730358) (50, 0.8524288884395915) (60, 0.8310530363592776) (70, 0.8126607315344698) (80, 0.7864716512974249) (90, 0.7545858840564603) (100, 0.7327127301131534) (110, 0.7091468609601421) (120, 0.6853617377346405) (130, 0.6635374727937213) (140, 0.6347498552392201) (150, 0.6046882392250652) (160, 0.5753770423023588) (170, 0.5373087863680107) (180, 0.5013835940062829) (190, 0.4675908469623859) (200, 0.4249509256398726) (210, 0.3786558813516064) (220, 0.31766671996795226) (230, 0.23809164865980076) (240, 0.16614377679904924) (250, 0.10430176579222451) (260, 0.06686376175016114) (270, 0.04699511013231828) (280, 0.037147257115501216) (290, 0.03128492306740041) (300, 0.026504127942343855) (310, 0.02266624374387016) (320, 0.019032414410667057) (330, 0.017291957147836757) (340, 0.015999717444953696) (350, 0.015069416884284537) (360, 0.014413853509388836) (370, 0.013788111993964076) (380, 0.011534694663308424) (390, 0.009691062357088918) (400, 0.008449305419103841) (410, 0.0073172665663327335) (420, 0.006896321907878315) (430, 0.006341017881434075) (440, 0.004251515330708114) (450, 0.0020691630588748713) (460, 0.0008829263790903801) (470, 0.0004978960002697326) (480, 0.000356328918336287) (490, 0.00014617131794757436) (491, 8.950505010408427e-05) (492, 6.347633162165342e-05) (493, 4.247892089181589e-05) (494, 2.8169379745910093e-05) (495, 1.7921748791707997e-05) (496, 1.159679616123573e-05) (497, 6.650472027143783e-06) (498, 4.416870070380276e-06) (499, 2.7622328159840534e-06) (500, 1.580449206628693e-06) (501, 5.297359306557617e-07) (502, 1.0009807687459418e-07) (503, 1.5020055897995587e-08) (504, 4.423869002155035e-09) (505, 2.7972359755621495e-11) 
    };
    \addlegendentry{Gauss Seidel}

    \addplot[mark=diamond,color=vir6] coordinates {
        (0, 1.0) (10, 0.9342163794989697) (20, 0.8714423424226219) (30, 0.810783217736397) (40, 0.7422212590634999) (50, 0.6669582173495461) (60, 0.5836208229833955) (70, 0.48691981378082905) (80, 0.36021125068379956) (90, 0.19517603665998032) (100, 0.08460897232104497) (110, 0.041958178668946265) (120, 0.02265203801477013) (130, 0.014959092278693048) (140, 0.012621022487225833) (150, 0.010377554392894789) (160, 0.0074892537188292925) (170, 0.005705879664209461) (180, 0.004018916376478654) (190, 0.0007789169596072272) (200, 3.0768956652096287e-09) (201, 1.549793282846289e-10) (202, 7.986057360694872e-11)  
    };
    \addlegendentry{Schwarz}
    
    \addplot[color=vir4, mark=square] coordinates {
        (0, 1.0) (10, 0.20987068243492948) (20, 0.1789714280141495) (30, 0.1360312958637759) (40, 0.12671180489474262) (50, 0.12213342336573228) (60, 0.10406408860105802) (70, 0.09738126690283455) (80, 0.09602748904971967) (90, 0.09039152271999154) (100, 0.08533781570338798) (110, 0.08394180932197401) (120, 0.08233358785924674) (130, 0.0800823523742912) (140, 0.07779115161176085) (150, 0.07639151731297043) (160, 0.07498133899449783) (170, 0.07420020998956083) (180, 0.07328176888535541) (190, 0.07221667688674273) (200, 0.06945063399782872) (210, 0.06854965067534728) (220, 0.06698588719279792) (230, 0.06550391830313694) (240, 0.06367406529014075) (250, 0.06286973411130088) (260, 0.06182142098572725) (270, 0.06094316692609111) (280, 0.05948790998272106) (290, 0.058283602272073486) (300, 0.05509523257240939) (310, 0.05372745658250891) (320, 0.04799001241894954) (330, 0.04679602360269621) (340, 0.04437638045589388) (350, 0.04109775296493586) (360, 0.03994146820039353) (370, 0.038495298003092884) (380, 0.03755602489567478) (390, 0.03578963335053881) (400, 0.0335000219505972) (410, 0.032516551295919134) (420, 0.014307139559427208) (430, 0.011434200834473293) (440, 0.009747828458090123) (450, 0.0044690646346497555) (460, 0.001363217520236999) (470, 0.00022013686993539497) (480, 6.357896065738972e-07) (489, 2.290390733974232e-09) (490, 1.2728558678777735e-09) (491, 5.832503034529717e-10) (492, 2.1390069181762254e-10) (493, 9.448112159260241e-11)  
    };
    \addlegendentry{UNet}
    
    \addplot[mark=+,color=vir5] coordinates {
        (0, 1.0) (10, 0.011730604492457964) (20, 0.009213665749090424) (30, 0.008935719253296295) (40, 0.008883718838718746) (50, 0.008832183290578518) (60, 0.008766019927134437) (70, 0.00818323373032284) (80, 0.0074261440463286674) (90, 0.007213294394461229) (100, 0.007171909516585396) (110, 0.007136444877588504) (120, 0.006914910949825328) (130, 0.006172441522542972) (140, 0.005799244134739346) (150, 0.005424950992920647) (160, 0.005254910928075411) (170, 0.005228571830080484) (180, 0.005224696685286327) (190, 0.00520848709350312) (200, 0.005028399196976466) (210, 0.0046091202338519365) (220, 0.004431201442220288) (230, 0.004344748623402378) (240, 0.004297647325397021) (250, 0.004294087498845729) (260, 0.004202281712686691) (270, 0.004115277635005418) (280, 0.00408101072631444) (290, 0.00342525546694761) (300, 0.001357229346521154) (310, 0.00034204131415973513) (320, 6.17123222475476e-05) (330, 1.3931036961002134e-05) (340, 3.133721203781303e-06) (350, 6.122537246031866e-07) (360, 1.1207963942771565e-07) (370, 1.761003084158037e-08) (380, 3.499498907288189e-09) (390, 6.533889145553497e-10) (396, 2.3053436322820832e-10) (397, 2.0434278589747513e-10) (398, 1.7940028513831587e-10) (399, 1.5767368692766794e-10) (400, 1.3261719150921442e-10) (401, 1.1268085967461527e-10) (402, 9.639075943217711e-11) 
    };
    \addlegendentry{FNO}

    \addplot[color=vir2, mark=pentagon] coordinates {
        (0, 1.0) (10, 0.009620116105752065) (20, 0.006642553764718631) (30, 0.0053012097150703565) (40, 0.0044977678789690665) (50, 0.003937793511855631) (60, 0.003504543460360238) (70, 0.003162279465077411) (80, 0.00287944580879148) (90, 0.0026354609442815354) (100, 0.0024234305781964424) (110, 0.0022368138318249616) (120, 0.0020670484464224573) (130, 0.0019116946191673422) (140, 0.0017675861245182536) (150, 0.0016317094638202147) (160, 0.001502777885146766) (170, 0.001380474303979282) (180, 0.001261354200973783) (190, 0.0011426823502089908) (200, 0.0010262876402775223) (210, 0.000907229642778926) (220, 0.0007819855550200552) (230, 0.0006493398839643633) (240, 0.0004997526304644904) (250, 0.00030059317719074486) (260, 4.5184691924434265e-09) (270, 2.3442959008810886e-09) (280, 1.660523356709187e-09) (290, 1.4015274054530073e-09) (300, 1.246063260229793e-09) (310, 1.1525346998388441e-09) (320, 1.084241268934601e-09) (330, 9.518725923860468e-10) (340, 8.590232121438264e-10) (350, 7.978529494937181e-10) (360, 7.648708441407299e-10) (370, 7.161775363417901e-10) (380, 6.356035534641201e-10) (390, 5.444267845233705e-10) (400, 4.850716738724034e-10) (410, 4.470756583873367e-10) (420, 4.1649971977949924e-10) (430, 3.7324677367209856e-10) (440, 3.360582750253797e-10) (450, 2.5727089477026975e-10) (460, 1.9390840265569493e-10) (466, 1.4312322691497687e-10) (467, 1.3707802560630945e-10) (468, 1.270397517564319e-10) (469, 1.1581370345940873e-10) (470, 1.0173869931377159e-10) (471, 9.026693721467517e-11) 
    };
    \addlegendentry{Transolver}
    
    \addplot[mark=triangle,color=vir0] coordinates {
        (0, 1.0) (10, 0.39656465216815406) (20, 0.09334475433911454) (30, 0.008303190921991417) (40, 0.0002470325129126541) (50, 3.1803738687405116e-08) (51, 9.113545842461378e-09) (52, 2.5088552672564824e-09) (53, 7.327896057090811e-10) (54, 2.093001218058249e-10) (55, 5.09654616254412e-11) 
    };
    \addlegendentry{M2NO}
    
\end{axis}
\end{tikzpicture}
\vspace{-10pt}
\caption{Preconditioner comparison with GMRES for solving Poisson equation.}
\label{fig:preconditioner}
\vspace{-10pt}
\end{figure}

\subsection{Model Analysis} 

\textbf{Efficiency Analysis.}
We evaluate the computational efficiency of M2NO against several neural operator baselines on a high-resolution input ($512\times512\times3$) with batch size 16 (see Table \ref{table:efficiency}). M2NO achieves superior performance-efficiency balance, having the \textbf{lowest parameter count (417k)} and modest GPU memory usage (17,992 MiB), significantly lighter than large-scale models such as MgNO (227M parameters) and LSM (4.8M parameters). Despite its small size, M2NO maintains competitive training (18.39 s/epoch) and inference (1.78 s/epoch) speeds. This highlights M2NO’s effectiveness for resource-constrained scenarios in scientific computing, delivering strong accuracy without compromising efficiency.

\subsection{Ablation Study Analysis}
We've performed additional ablation studies on the Navier–Stokes dataset (Table\ref{table:ablation}) to further clarify each component's contribution to M2NO. The results clearly indicate that removing either the multilevel, multigrid, or multi-wavelet component leads to a noticeable increase in the $L_2$ error, demonstrating the necessity of each for optimal performance. Additionally, substituting the multi-wavelet transform with FFT achieves competitive results (3.97E-03), confirming our framework’s overall robustness. However, accuracy notably decreases compared to our multi-wavelet baseline (3.61E-03). This emphasizes the importance of the spatial localization capability provided by multiwavelets, which is crucial for accurately capturing localized, multiscale PDE features.

\begin{table}[h]
\centering
\caption{Ablation results of different M2NO configurations.}
\vspace{-5pt}
\label{table:ablation}
\begin{sc}
\resizebox{0.8\linewidth}{!}{
\begin{tabular}{l|l}
\toprule
Model Configuration & $L_{2}$ Error \\ 
\midrule
w/o Multilevel & 4.47E-03 \\
w/o Multigrid & 4.55E-03 \\
w/o Multi-wavelet & 4.36E-03 \\
FFT as Mapping Operator & 3.97E-03 \\
\midrule
M2NO (Baseline) & 3.61E-03 \\ 
\bottomrule
\end{tabular}}
\end{sc}
\end{table}

\subsection{Hyperparameter Study Analysis}

The hyperparameter study of M2NO, presented in Table \ref{table:hyper}, explored the impact of various model configurations on the $L_2$ error. Adjustments were made across multiple parameters, including iteration steps, grid levels, wavelet numbers, basis numbers, and the number of layers. Notably, increasing the grid levels from a flat [1,1,1,1,1,1] to a varied [1,2,3] configuration reduced the $L_2$ error from 7.35E-06 to 6.45E-06. Changes in the wavelet number revealed that using c=16 minimized the error to 5.29E-06, indicating a more significant wavelet count enhances accuracy. Similarly, an optimal layer count was identified, with six layers achieving the lowest error at 4.66E-06. Conversely, varying the basis number showed that a higher count does not always correlate with lower error, as k=4 performed better than k=2. These results illustrate the nuanced impacts of each parameter on model performance, guiding further refinement and optimization of M2NO configurations for enhanced precision in scientific simulations.

\begin{table}[h]
\centering
\caption{Hyperparameter study results for the M2NO model.}
\vspace{-5pt}
\label{table:hyper}
\begin{sc}
\resizebox{\linewidth}{!}{
\begin{tabular}{c|ll}
\toprule
Hyperparameter Type             & Model   Configuration & $L_{2}$ Error \\ 
\midrule
Iteration                       & {[}1,1,1,1,1,1,1{]}   & 7.35E-06 \\ 
\midrule
Grid Level                      & {[}1,2,3{]}           & 6.45E-06 \\ 
\midrule
\multirow{4}{*}{Wavelet Number} & c=2                   & 1.83E-05 \\
                                & c=4                   & 8.16E-06 \\
                                & c=6                   & 6.20E-06 \\
                                & c=16                  & 5.29E-06 \\ 
\midrule
\multirow{2}{*}{Basis Number}   & k=2                   & 1.28E-05 \\
                                & k=4                   & 7.15E-06 \\ 
\midrule
\multirow{3}{*}{Layer Number}   & layer=3               & 8.24E-06 \\
                                & layer=4               & 7.19E-06 \\
                                & layer=6               & 4.66E-06 \\ 
\midrule
Baseline                        & M2NO                  & 7.16E-06 \\ 
\bottomrule
\end{tabular}}
\end{sc}
\end{table}

\section{Related Work}

Recent advances in neural operators have explored multigrid and wavelet-based architectures to improve efficiency and multi-scale modeling for PDEs. Below, we compare M2NO with representative multigrid and wavelet neural operator approaches.

\textbf{Multigrid Neural Operators.} MgNO~\cite{23:mgno} applies multigrid hierarchies to neural operators for efficient linear operator parameterization, but it lacks multiresolution analysis and is limited to linear PDEs. Methods like UGrid~\cite{24:UGrid} and Hsieh et al.~\cite{19:LearningNP} similarly accelerate linear solvers via multigrid-inspired neural modules with convergence guarantees, yet remain constrained to linear regimes without explicit multi-scale error correction. In contrast, M2NO integrates multiwavelet-based restriction and prolongation within a learnable multigrid cycle, enabling coarse-to-fine error correction for nonlinear, multiscale PDEs.

\textbf{Wavelet-based Neural Operators.} MWT~\cite{21:mwt} and its variants, including CMWOL~\cite{23:CoupledMO} and WDNO~\cite{25:WaveletDNO}, leverage multiwavelet transforms to capture localized and multi-scale features, but do not incorporate an explicit hierarchical multigrid structure for iterative error propagation. M2NO advances beyond these by unifying multiwavelet representation with a full multigrid correction framework, supporting efficient error management across all scales.

Technically, M2NO distinguishes itself by combining multiwavelet-based multiresolution analysis with an explicit, learnable multigrid hierarchy. This enables robust coarse-to-fine error correction for nonlinear and multiscale PDEs, bridging gaps left by prior multigrid and wavelet neural operator methods.

\section{Conclusion}\label{sec:con}

We introduced M2NO, a \textbf{multigrid} neural operator whose restriction and prolongation
steps use pre\-defined \textbf{multiwavelet} filters.  
This hybrid architecture (i) captures fine–to–coarse interactions in a unified multi-scale representation,  
(ii) achieves state-of-the-art accuracy with low parameter cost on diverse 1-D/2-D PDE benchmarks (including ERA5),  
and (iii) can serve as an efficient preconditioner for classical iterative solvers.

\textbf{Limitation.} (i) M2NO currently assumes structured grids; accuracy degrades on highly non-uniform or adaptive meshes.  
(ii) Very deep multigrid cycles may incur communication overhead on distributed hardware.  

\textbf{Future work.} We will (i) extend M2NO to unstructured and adaptive grids by coupling graph-based wavelet kernels with geometric multigrid, (ii) develop dynamic level-of-detail schedules that adjust wavelet depth online, and (iii) systematically investigate its preconditioning power in large-scale scientific simulations. These directions aim to broaden M2NO’s applicability to a wider class of scientific and engineering problems while further improving computational efficiency.

\begin{acks}
    Wei Wang was supported by Advanced Materials-National Science and Technology Major Project (Grant No. 2025ZD0620100), and the Key Research and Development Program in Sichuan Province of China (No. 2024YFFK0410). 
    Zhilu Lai was supported by the Guangdong Provincial Fund - Special Innovation Project (2024KTSCX038) and Research Grants Council of Hong Kong through the Research Impact Fund (R5006-23). 
    Xiaobo Zhang was supported by the Key Research and Development Program in Sichuan Province of China (No. 2024YFFK0410).
\end{acks}

\bibliographystyle{ACM-Reference-Format}
\bibliography{reference}

\appendix

\section{Multigrid Methods} \label{appendix:amg}

The key ingredients are the operators $I_{h}^{2h}$ and $I_{2h}^{h}$ that change grids:
\begin{itemize}
    \item A restriction operator $I_{h}^{2h}$ transfers vectors from the fine grid to the coarse grid.
    \item An interpolation operator $I_{2h}^{h}$ returns to the fine grid.
    \item The original matrix $A_{h}$ on the fine grid is approximated by $A_{2h}=I_{h}^{2h}A_{h}I$ on the coarse grid.
\end{itemize}

The spectral radius $\rho(I_{h}^{2h})$ is called the convergence factor. It is roughly the worst factor by which the error is reduced with each relaxation sweep.

\subsection{Multigrid Theory}

Recall the variational properties:
\begin{equation}
    \begin{aligned}
        A^{2h}&=I^{2h}_{h}A^{h}I^{h}_{2h}&(\text{Galerkin property}),\\
        I^{2h}_{h}&=c(I^{h}_{2h})^{T},&c\in I_{h}^{2h}.
    \end{aligned}
\end{equation}

From the orthogonality relationships between the subspaces of a linear operator (Fundamental Theorem of Linear Algebra), we know that
\begin{equation}
    \text{Nullspace}(I^{2h}_{h})\;\bot \;\text{Range}[(I^{2h}_{h})^{T}].
\end{equation}

By the second variational property, it then follows that:
\begin{equation}
    \text{Nullspace}(I^{2h}_{h})\;\bot\; \text{Range}(I^{h}_{2h}),
\end{equation}
which means that $(\mathbf{q}^{h},\mathbf{u}^{h})=0$ whenever $\mathbf{q}^{h}\in \text{Range}(I_{2h}^{h})$ and $I_{h}^{2h}\mathbf{u}^{h}=0$. This is equivalent to the condition that $(\mathbf{q}^{h},A^{h}\mathbf{u}^{h})=0$ whenever $\mathbf{q}^{h}\in \text{Range}(I_{2h}^{h})$ and $I_{h}^{2h}A^{h}\mathbf{u}^{h}=0$.

This orthogonality property allows the space $\Omega^{h}$ to be decomposed in the form:
\begin{equation}
    \Omega^{h}=\text{Range}(I_{2h}^{h})\oplus \text{Nullspace}(I_{h}^{2h}A^{h}),
\end{equation}
which means that if $\mathbf{e}^{h}$ is a vector  in $\Omega^{h}$, then it may always be expressed as 
\begin{equation}
    \mathbf{e}^{h}=\mathbf{s}^{h}+\mathbf{t}^{h},
\end{equation}
where $\mathbf{s}^{h}\in \text{Range}(I_{2h}^{h})$ and $\mathbf{t}^{h}\in \text{Nullspace}(I_{h}^{2h}A^{h})$. We associate $\mathbf{s}^{h}$ with the smooth components of the error and we associate $\mathbf{t}^{h}$ with the oscillatory components of the error.

\RestyleAlgo{ruled}
\begin{algorithm}
\caption{V-cycle}
\label{alg:v}
\SetKwInOut{Input}{Input}
\SetKwInOut{Output}{Output}

\Input{Grid level $\Omega^h$ with operator $A^h$; initial guess $v^h$; right-hand side $f^h$; pre-smoothing steps $\alpha_1$; post-smoothing steps $\alpha_2$.}
\Output{Updated approximation $v^h$.}

\If{$\Omega^h$ is the coarsest grid}{
    Relax $\alpha_2$ times on $A^h u^h = f^h$ with initial guess $v^h$\;
}
\Else{
    Relax $\alpha_1$ times on $A^h u^h = f^h$ with initial guess $v^h$\;
    $r^h \gets f^h - A^h v^h$\tcp*{Compute the residual}
    $f^{2h} \gets I^{2h}_h r^h$\tcp*{Restrict the residual to the coarser grid}
    $v^{2h} \gets 0$\tcp*{Initial guess for the coarser grid}
    $v^{2h} \gets \text{Vcycle}(v^{2h}, f^{2h})$\tcp*{Recursive call}
    $v^h \gets v^h + I^h_{2h} v^{2h}$\tcp*{Prolongate the correction and update}
    Relax $\alpha_2$ times on $A^h u^h = f^h$ with initial guess $v^h$\;
}

\Return{$v^h$}\;
\end{algorithm}

\section{Multiwavelets} \label{appendix:mw}

\begin{figure*}[t]
    \centering
    \subfigure[The two-dimensional discrete wavelet transform.]{\includegraphics[width=0.58\linewidth]{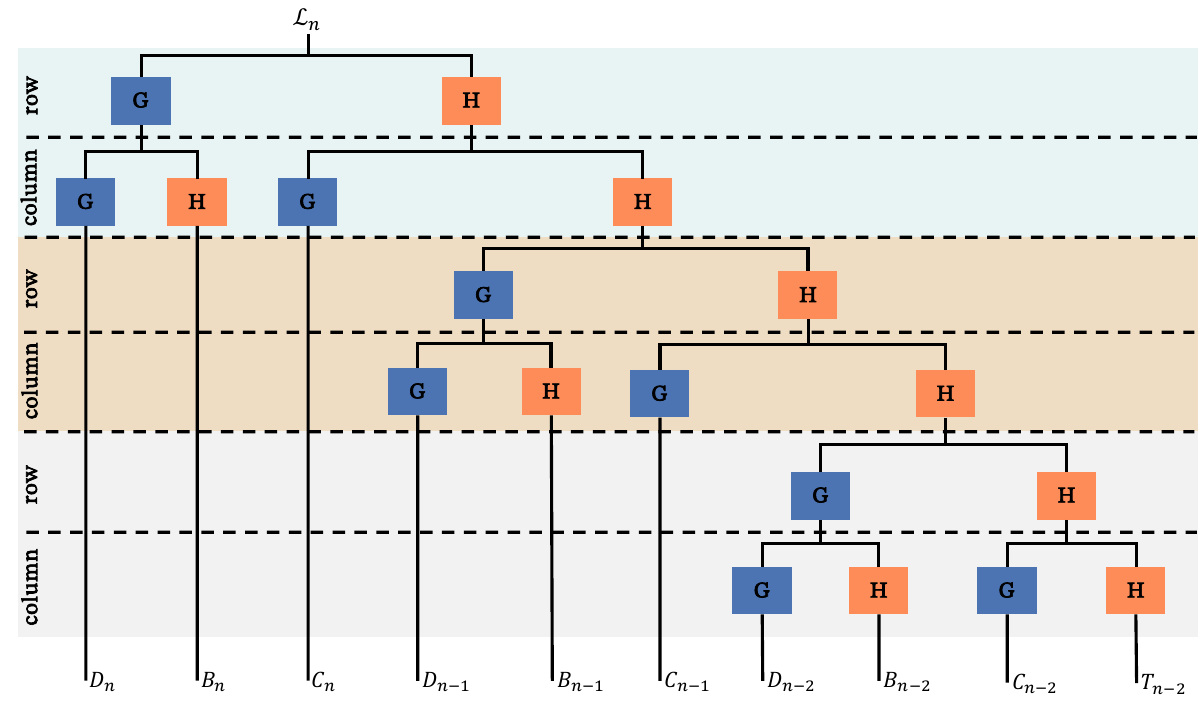}}
    \subfigure[Matrix representation of decomposition.]{\includegraphics[width=0.4\linewidth]{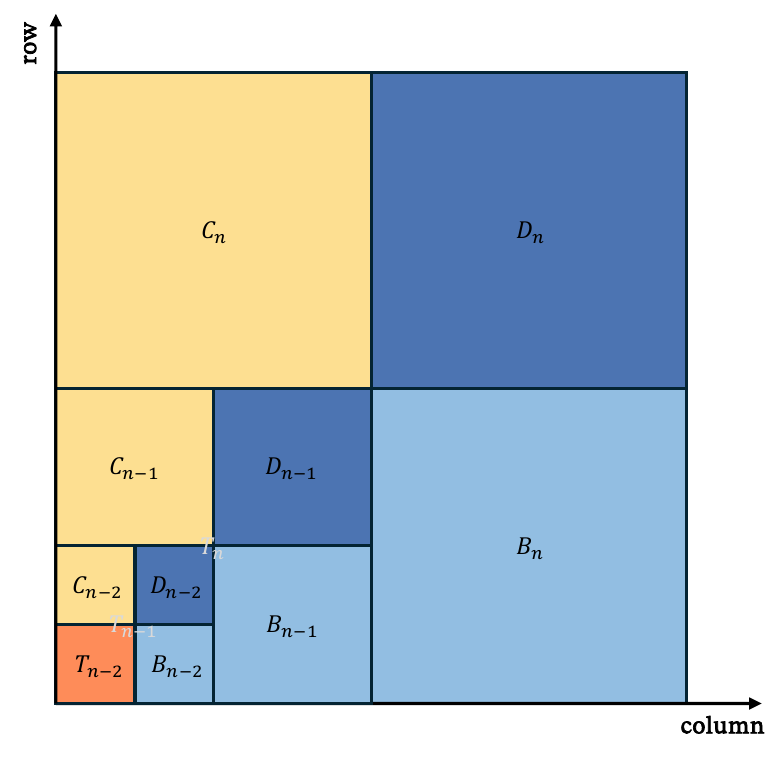}}
    \caption{(a) Two-dimensional discrete wavelet transform with $G$ (high-pass) and $H$ (low-pass) filters. Horizontal lines represent frequency bands; vertical lines show decomposition levels. (b) Matrix representation of decomposition, dividing \( L_n \) into submatrices \( B_n \), \( C_n \), and \( D_n \).}
    \label{fig:dwt}
\end{figure*}

\subsection{Multiresolution analysis}
For $k=1,2,\ldots$, and $n=0,1,2,\ldots$, we define $\mathbf{V}_{n}^{k}$ as a space of piecewise polynomial functions, 
\begin{equation}
\begin{aligned}
    \mathbf{V}_{n}^{k}=\{&f:\text{the restriction of }f\text{ to the interval }(2^{-n}l,2^{-n}(l+1))\\ &\text{ is a polynomial} \\
    &\text{of degree less than }k\text{, for }l=0,\ldots,2^{n}-1 \\ &\text{, and }f\text{ vanishes elsewhere}\}.
\end{aligned}
\end{equation}

The space $\mathbf{V}_{n}^{k}$ has dimension $2^{n}k$ and
\begin{equation} \label{mw-0}
    \mathbf{V}_{0}^{k}\subset\mathbf{V}_{1}^{k}\subset\ldots\subset\mathbf{V}_{n}^{k}\subset\ldots.
\end{equation}

We define the multiwavelet subspace $\mathbf{W}_{n}^{k}$, $n=0,1,2,\ldots$, as the orthogonal complement of $\mathbf{V}_{n}^{k}$ in $\mathbf{V}_{n+1}^{k}$,
\begin{equation} \label{mw-1}
    \mathbf{V}_{n}^{k}\oplus\mathbf{W}_{n}^{k}=\mathbf{V}_{n+1}^{k},\quad \mathbf{W}_{n}^{k}\bot\mathbf{V}_{n}^{k},
\end{equation}
and note that $\mathbf{W}_{n}^{k}$ is of dimension $2^{n}k$. Therefore, we have
\begin{equation}
    \mathbf{V}_{n}^{k}=\mathbf{V}_{0}^{k}\oplus\mathbf{W}_{0}^{k}\oplus\mathbf{W}_{1}^{k}\oplus\ldots\oplus\mathbf{W}_{n-1}^{k}.
\end{equation}

We define $\mathbf{V}^{k}=\bigcup\limits_{n=0}^{\infty}{V_{n}^{k}}$ and observe that $\mathbf{V}^{k}$ is dense in $L^{2}([0,1])$ with respect to norm $\Vert{f}\Vert=\langle{f,f}\rangle^{1/2}$, where
\begin{equation} \label{mw-4}
    \langle{f,g}\rangle=\int_{0}^{1}{f(x)g(x)dx}.
\end{equation}

Given a basis $\phi_{0},\ldots,\phi_{k-1}$ of $\mathbf{V}_{0}^{k}$, the space $\mathbf{V}_{n}^{k}$ is spanned by $2^{n}k$ functions which are obtained from $\phi_{0},\ldots,\phi_{k-1}$ by dilation and translation,
\begin{equation}
    \phi_{jl}^{n}=2^{n/2}\phi_{j}(2^{n}x-l),\quad j=0,\ldots,k-1,\quad l=0,\ldots,2^{n}-1.
\end{equation}

\subsection{Multiwavelets}
We introduce piecewise polynomial functions $\psi_{0}\ldots,\psi_{k-1}$ to be an orthonormal basis for $\mathbf{W}_{0}^{k}$,
\begin{equation}
    \int_{0}^{1}{\psi_{i}(x)\psi_{j}(x)dx}=\delta_{ij}.
\end{equation}
Since $\mathbf{W}_{0}^{k}\bot\mathbf{V}_{0}^{k}$, the first $k$ moments of $\psi_{0},\ldots,\psi_{k-1}$ vanish:
\begin{equation}
    \int_{0}^{1}{\psi_{j}(x)x^{i}dx}=0,\quad i,j=0,1,\ldots,k-1.
\end{equation}

The space $\mathbf{W}_{n}^{k}$ is spanned by $2^{n}k$ functions obtained from $\psi_{0},\ldots,\psi_{k-1}$ by dilation and translation,
\begin{equation}
    \psi_{jl}^{n}(x)=2^{n/2}\psi_{j}(2^{n}x-l),\quad j=0,\ldots,k-1,\quad l=0,\ldots,2^{n}-1,
\end{equation}
and supp $(\psi_{jl}^{n})=I_{nl}$ denotes the interval $[2^{-n}l,2^{-n}(l+1)]$. The condition of orthonormality of $\psi_{0},\ldots,\psi_{k-1}$ yields
\begin{equation}
    \int_{0}^{1}{\psi_{il}^{n}(x)\psi_{jm}^{n'}(x)dx}=\delta_{ij}\delta_{lm}\delta_{nn'}.
\end{equation}
The set $\{\phi_{0},\ldots,\phi_{k-1}\}\cup\{\psi_{jl}^{n}:j=0,\ldots,k-1,l=0,\ldots,2^{n}-1,n=0,1,\ldots\}$ therefore forms a complete orthonormal basis for $\mathbb{L}^{2}([0,1])$.

\subsection{Multiwavelet Transformation}
In order to compute projections of functions on subspaces of multiresolution analysis, we consider the two-scale difference equations. In our case, functions $\phi$ and $\psi$ are known, and we use them to construct the filter coefficients. The relations \eqref{mw-0} and \eqref{mw-1} between the subspaces may be expressed by the two-scale difference equations,
\begin{equation} \label{mw-2}
    \phi_{i}(x)=\sqrt{2}\sum_{j=0}^{k-1}{(h_{ij}^{(0)}\phi_{j}(2x)-h_{ij}^{(1)}\phi_{j}(2x-1))},\quad i=0,\ldots,k-1,
\end{equation}
\begin{equation} \label{mw-3}
    \psi_{i}(x)=\sqrt{2}\sum_{j=0}^{k-1}{(g_{ij}^{(0)}\phi_{j}(2x)+g_{ij}^{(1)}\phi_{j}(2x-1))},\quad i=0,\ldots,k-1,
\end{equation}
where the coefficients $g_{ij}^{(0)}$, $g_{ij}^{(1)}$ depend on the choice of the order $k$. The function $\sqrt{2}\phi_{0}(2x)\ldots,\sqrt{2}\phi_{k-1}(2x)$ in \eqref{mw-2} are orthonormal on the interval $[0,\frac{1}{2}]$ whereas $\sqrt{2}\phi_{0}(2x-1)\ldots,\sqrt{2}\phi_{k-1}(2x-1)$ are orthonormal on the interval $[\frac{1}{2},1]$. The matrices of coefficients
\begin{equation}
    H^{(0)}=\{h_{ij}^{(0)}\},\quad H^{(1)}=\{h_{ij}^{(1)}\},\quad G^{(0)}=\{g_{ij}^{(0)}\},\quad G^{(1)}=\{g_{ij}^{(1)}\}
\end{equation}
are analogs of the quadrature mirror filters. The two-scale equations \eqref{mw-2},\eqref{mw-3} lead us to a multiresolution decomposition. 

By construction, we have $\langle{\phi_{i},\phi_{j}}\rangle=\delta_{ij}$, $\langle{\psi_{i},\psi_{j}}\rangle=\delta_{ij}$, and $\langle{\phi_{i},\phi_{j}}\rangle=0$ for $i,j=0,\ldots,k-1$, where $\langle,\rangle$ is the inner product defined in \eqref{mw-4}. Using this orthogonality conditions and (\ref{mw-2}-\ref{mw-3}), we obtain
\begin{align}
    H^{(0)}H^{(0)T}+H^{(1)}H^{(1)T}&=I,\\
    G^{(0)}G^{(0)T}+G^{(1)}G^{(1)T}&=I,\\
    H^{(0)}G^{(0)T}+H^{(1)}G^{(1)T}&=0.
\end{align}

The matrices of coefficients $H^{(0)}, H^{(1)}, G^{(0)}, G^{(1)}$ are analogs of the quadrature mirror filter. We now derive the necessary relations for multiresolution reconstruction. In one dimension, we construct
\begin{equation}
    \begin{aligned}
        H&=\begin{pmatrix}
            H^{(0)}&H^{(1)}
        \end{pmatrix}\\
        G&=\begin{pmatrix}
            G^{(0)}&G^{(1)}
        \end{pmatrix}
    \end{aligned}
\end{equation}
 to be the operators that transform the basis of the space $V_{j}$ to the bases of the spaces $V_{j+1}$ and $W_{j+1}$, respectively. And $H$ and $G$ play the role of separating the low and high frequency components. With the property of $H^{(0)}, H^{(1)}, G^{(0)}, G^{(1)}$, we have
\begin{equation}
    \begin{aligned}
        H^{T}H+G^{T}G&=I\\
        HG^{T}=GH^{T}&=0\\
        HH^{T}=GG^{T}&=I
    \end{aligned}
\end{equation}
In two dimensions, the tensor product of one-dimensional multiresolution analyses is used. Analogous to the one-dimensional case, define the operators $H$ and $G$ so that
\begin{equation}
    \begin{aligned}
        H&=H_{y}\otimes H_{x}\\
        G&=\begin{pmatrix}
            G_{y}\otimes H_{x}\\ H_{y}\otimes G_{x}\\ G_{y}\otimes G_{x}
        \end{pmatrix},
    \end{aligned}
\end{equation}
where $\otimes$ is the Kronecker product.

We denote the multiwavelet transform as
\begin{equation}
    \mathcal{W}=\begin{pmatrix}
        H\\G
    \end{pmatrix},
\end{equation}
Due to the properties of $H$ and $G$ , $\mathcal{W}$ is orthogonal.

\section{Proof of Theorem \ref{th:conv}}
\label{appendix:proof}

We show that the \emph{Multi\-wavelet–Based Multigrid Neural Operator} (M2NO) can approximate any continuous operator
$O^{*}:X\!\to\!Y$ to arbitrary accuracy, where
$X = H^{s}(\Omega)$, $Y = H^{s'}(\Omega)$ for $s,s' \ge 1$, and
$\Omega\subset\mathbb R^{d}$ is a bounded Lipschitz domain.
Throughout, $\|\!\cdot\!\|_{X}$ and $\|\!\cdot\!\|_{Y}$ denote the
respective Sobolev norms, and $L=\|O^{*}\|_{\mathcal L(X,Y)}$.

\subsection*{Step\,1:\;Finite–Element Projection (C.1)}
Because piecewise–polynomial spaces are dense in $H^{s}(\Omega)$,
for every $\varepsilon>0$ there exists a mesh parameter $h>0$ and a
finite–dimensional subspace
$V_{h}=\operatorname{span}\{\psi_{1},\dots,\psi_{k}\}\subset X$
such that
\begin{equation}\label{eq:fem-approx}
  \sup_{u\in C}\inf_{v_{h}\in V_{h}}\|u-v_{h}\|_{X}
  \;<\;\frac{\varepsilon}{3L},
\end{equation}
where $C\subset X$ is any compact set of interest
(e.g.\ the training support).  
Let $\Pi_{h}\!:X\!\to\!V_{h}$ be the
$X$–orthogonal projection.  Standard finite–element theory
\cite{13:fem} then yields
\begin{equation}\label{eq:fem-bound}
  \|O^{*}(u)-O^{*}(\Pi_{h}u)\|_{Y}
  \;\le\;L\|u-\Pi_{h}u\|_{X}
  \;<\;\frac{\varepsilon}{3}\quad\forall\,u\in C.
\end{equation}

\subsection*{Step\,2:\;Operator Discretisation (C.2)}
Because $O^{*}(V_{h})\subset Y$ lies in a finite–dimensional
subspace, there exist basis functions
$\{\phi_{1},\dots,\phi_{m}\}\subset Y$
and continuous coefficient maps
$f_{i}\!:V_{h}\!\to\!\mathbb R$
such that, for every $v_{h}\in V_{h}$,
\begin{equation}\label{eq:basis-approx}
  \Bigl\|O^{*}(v_{h})-\sum_{i=1}^{m}f_{i}(v_{h})\,\phi_{i}\Bigr\|_{Y}
  \;<\;\frac{\varepsilon}{3}.
\end{equation}
Inequality \eqref{eq:basis-approx} follows from
the Weierstrass approximation theorem applied on the compact set
$O^{*}\bigl(C\bigr)$.

\subsection*{Step\,3:\;M2NO Construction (C.3)}
M2NO realises a map
$\widehat{O}_{n}\!:V_{h}\!\to\!\operatorname{span}\{\phi_{i}\}_{i=1}^{m}$
by three modules:
(i) projection $u\mapsto\Pi_{h}u$;
(ii) multiwavelet restriction/prolongation, giving a
multi–resolution representation of the coefficient vector
$x=\bigl(\langle\Pi_{h}u,\psi_{j}\rangle\bigr)_{j=1}^{k}$; and
(iii) a learnable linear map
$A_{n}=[a_{ij}]_{i=1,\dots,m}^{j=1,\dots,k}$ trained to minimise
\[
  E_{n}\;=\;\sup_{x\in\mathbb R^{k},\,\|x\|\le 1}
  \Bigl\|\sum_{j=1}^{k}x_{j}\psi_{j}-\widehat{O}_{n}\!\!\left(
        \sum_{j=1}^{k}x_{j}\psi_{j}\right)\Bigr\|_{Y}.
\]
Universal approximation of linear maps by neural networks
ensures that we can choose $n$ large enough so that
$E_{n}<\varepsilon/3$.  
Equivalently,
\begin{equation}\label{eq:learnable-bound}
  \sup_{v_{h}\in V_{h},\,\|v_{h}\|_{X}\le 1}
  \|O^{*}(v_{h})-\widehat{O}_{n}(v_{h})\|_{Y}
  \;<\;\frac{\varepsilon}{3}.
\end{equation}

\subsection*{Step\,4:\;Wavelet Error Decomposition (Eq.\,50)}
Error residuals between $O^{*}$ and $\widehat{O}_{n}$ decompose into
wavelet scales:
\[
  O^{*}(u)-\widehat{O}_{n}(u)
  \;=\;r_{0}\;+\;\sum_{\ell=1}^{L}r_{\ell},
\]
where $r_{0}=O^{*}(u)-O^{*}(\Pi_{h}u)$ and
$r_{\ell}$ is the contribution of level–$\ell$ wavelets in the multigrid
cycle.  
Orthogonality of the multiwavelet basis
implies $\|r_{0}+\dots+r_{L}\|_{Y}^{2}
       =\|r_{0}\|_{Y}^{2}+\sum_{\ell=1}^{L}\|r_{\ell}\|_{Y}^{2}$,
allowing each term to be bounded independently by
\eqref{eq:fem-bound} and \eqref{eq:learnable-bound}.  Summing yields
\begin{equation}\label{eq:final-sup}
  \sup_{u\in C}\|O^{*}(u)-\widehat{O}_{n}(u)\|_{Y}
  \;\le\;\frac{\varepsilon}{3}
         \;+\;\frac{\varepsilon}{3}
         \;+\;\frac{\varepsilon}{3}
  \;=\;\varepsilon.
\end{equation}

\paragraph{Conclusion.}
For any $\varepsilon>0$, there exists a mesh size $h$,
wavelet depth $L$, and network width $n$ such that
\eqref{eq:final-sup} holds.  Hence M2NO is a universal approximator
for continuous operators $O^{*}\!:H^{s}(\Omega)\!\to\!H^{s'}(\Omega)$.

\end{document}